\documentclass[conference,a4paper]{APSIPA2021}
\usepackage{multirow}
\usepackage{graphicx}
\usepackage[font=footnotesize,justification=centering]{caption}
\usepackage[font=footnotesize,justification=centering]{subcaption}
\usepackage{amsmath}
\usepackage{amssymb}
\usepackage{amsxtra}
\usepackage[colorlinks = true,
            linkcolor = blue,
            urlcolor  = blue,
            citecolor = blue,
            anchorcolor = blue]{hyperref}
\newcommand{\MYhref}[3][blue]{\href{#2}{\color{#1}{#3}}}%
            
\usepackage{algorithm}% http://ctan.org/pkg/algorithm
\usepackage{algpseudocode}% http://ctan.org/pkg/algorithmicx

\usepackage{etoolbox}
\usepackage{tikz}
\usetikzlibrary{tikzmark}
\usetikzlibrary{calc}

\errorcontextlines\maxdimen

\newcommand{\ALGtikzmarkcolor}{black}% customise this, if you want
\newcommand{\ALGtikzmarkextraindent}{4pt}% customise this, if you want
\newcommand{\ALGtikzmarkverticaloffsetstart}{-.5ex}% customise this, if you want
\newcommand{\ALGtikzmarkverticaloffsetend}{-.5ex}% customise this, if you want
\makeatletter
\newcounter{ALG@tikzmark@tempcnta}

\newcommand\ALG@tikzmark@start{%
    \global\let\ALG@tikzmark@last\ALG@tikzmark@starttext%
    \expandafter\edef\csname ALG@tikzmark@\theALG@nested\endcsname{\theALG@tikzmark@tempcnta}%
    \tikzmark{ALG@tikzmark@start@\csname ALG@tikzmark@\theALG@nested\endcsname}%
    \addtocounter{ALG@tikzmark@tempcnta}{1}%
}

\def\ALG@tikzmark@starttext{start}
\newcommand\ALG@tikzmark@end{%
    \ifx\ALG@tikzmark@last\ALG@tikzmark@starttext
    \else
        \tikzmark{ALG@tikzmark@end@\csname ALG@tikzmark@\theALG@nested\endcsname}%
        \tikz[overlay,remember picture] \draw[\ALGtikzmarkcolor] let \p{S}=($(pic cs:ALG@tikzmark@start@\csname ALG@tikzmark@\theALG@nested\endcsname)+(\ALGtikzmarkextraindent,\ALGtikzmarkverticaloffsetstart)$), \p{E}=($(pic cs:ALG@tikzmark@end@\csname ALG@tikzmark@\theALG@nested\endcsname)+(\ALGtikzmarkextraindent,\ALGtikzmarkverticaloffsetend)$) in (\x{S},\y{S})--(\x{S},\y{E});%
    \fi
    \gdef\ALG@tikzmark@last{end}%
}

\apptocmd{\ALG@beginblock}{\ALG@tikzmark@start}{}{\errmessage{failed to patch}}
\pretocmd{\ALG@endblock}{\ALG@tikzmark@end}{}{\errmessage{failed to patch}}
\makeatother
\usepackage{threeparttable}
\newcommand{\bS}[1]{{\boldsymbol{#1}}}

\long\def\comment#1{}

\newfont{\bbb}{msbm10 scaled 700}

\newfont{\bb}{msbm10 scaled 1000}

\algnewcommand{\LeftComment}[1]{\Statex \(\triangleright\) #1}

\begin{document}

\title{Channel-Wise Early Stopping without a Validation Set via NNK Polytope Interpolation}
% \title{CW-NNK Learning: Early Stopping without a Validation set via Channel-Wise Model Analysis}
%\title{Channel-Wise Early Stopping Without a Validation Set based on Polytope Interpolation}
% \title{Improved Neural Network Generalization using Channel-Wise NNK Graphs}
% Channel-wise NNK early stopping: improving neural network generalization without a validation set

\author{%
\authorblockN{%
David Bonet\authorrefmark{1}, Antonio Ortega\authorrefmark{2}, Javier Ruiz-Hidalgo\authorrefmark{1},
Sarath Shekkizhar\authorrefmark{2}
}
\authorblockA{%
\authorrefmark{1}
Universitat Politècnica de Catalunya, Barcelona, Spain}
\authorblockA{%
\authorrefmark{2}
University of Southern California, Los Angeles, USA}
\authorblockA{
email: \MYhref[black]{mailto:davidbonetsole@gmail.com}{davidbonetsole@gmail.com}, \MYhref[black]{mailto:aortega@usc.edu}{aortega@usc.edu}, \MYhref[black]{mailto:j.ruiz@upc.edu}{j.ruiz@upc.edu}, \MYhref[black]{mailto:shekkizh@usc.edu}{shekkizh@usc.edu}}
}

\maketitle
\thispagestyle{empty}

\begin{abstract}
  State-of-the-art neural network architectures continue to scale in size and deliver impressive generalization results, although this comes at the expense of limited interpretability. In particular, a key challenge is to determine when to stop training the model, as this has a significant impact on generalization. Convolutional neural networks (ConvNets) comprise high-dimensional feature spaces formed by the aggregation of multiple channels, where analyzing intermediate data representations and the model's evolution can be challenging owing to the curse of dimensionality. We present channel-wise DeepNNK (CW-DeepNNK), a novel channel-wise generalization estimate based on non-negative kernel regression (NNK) graphs with which we perform local polytope interpolation on low-dimensional channels. This method leads to instance-based interpretability of both the learned data representations and the relationship between channels. Motivated by our observations, we use CW-DeepNNK to propose a novel early stopping criterion that (i) does not require a validation set, (ii) is based on a task performance metric, and (iii) allows stopping to be reached at different points for each channel. Our experiments demonstrate that our proposed method has advantages as compared to the standard criterion based on validation set performance.
\end{abstract}

%\begin{IEEEkeywords}
%Convolutional networks, early stopping, generalization, channel-wise graph analysis, polytope interpolation.
%\end{IEEEkeywords}

\section{Introduction}

Graphs play an important role in many machine learning applications and are used to model data structures and similarities in a dataset \cite{ortega2018graph,ortega_2021}.
% Powerful properties, such as their compositional nature and 
The ability of graphs to define relationships between different types of entities allows us to describe and analyze complex patterns in data \cite{bacciu2020gentle}. 
%, e.g. chemical compounds, social networks, buying behaviours or 3D point clouds. 
% This and the increasing availability of computational resources has motivated a recent surge in interest in processing graphs with deep learning models. Graph Neural Networks \cite{scarselli2008graph} and recently Graph Convolutional (Neural) Networks \cite{kipf2016semi} are some of the predominant models that have overcome the big challenges of processing graphs in an adaptive fashion, including expressiveness and computational complexity, compared to the standard processing with vectorial data. 
%There is also a growing variety of graph-based tools for machine learning, that are used  such as neural networks. 
Recently, graphs have been used to understand and improve intermediate representations of deep learning models with application to various tasks, such as model regularization \cite{lassance2021laplacian} and robustness \cite{lassance2021representing}, model distillation \cite{lassance2020deep}, and model interpretation \cite{shekkizhar2020deepnnk, gripon2018inside}.
% , which is the focus of this work.
Since no graph is given a priori, these methods typically begin with a graph construction phase, where each graph node corresponds to an item in the training set, and the weight of an edge between two nodes is a function of the distance between their respective intermediate layer activations (i.e., their respective feature vectors).
% A standard way of applying graphs in neural networks involves using intermediate layer representations as feature vectors.
% This poses a problem in terms of runtime and memory consumption as one often encounters high dimensional feature representations, for example outputs of Convolutional Neural Networks (ConvNets) \cite{zagoruyko2016wide, tan2019efficientnet}. 
However, as deep learning models continue to grow in size, it is unclear if graphs constructed  in these increasingly higher dimension spaces \cite{papernot2018, shekkizhar2020} are able to capture relevant similarity information owing to the \textit{curse of dimensionality}. Further, higher dimensions also increase the computational requirements for graph construction.

In this work, we use a novel graph construction and develop an improved model analysis for deep learning by taking advantage of a property of convolutional neural networks (ConvNets) \cite{he2016deep, zagoruyko2016wide, tan2019efficientnet}.
%, namely, their use of multiple filters at each layer. 
Specifically, we note that convolutional layers (generally followed by non-linear and pooling layers) are formed by multiple convolutional filters that are applied in parallel to a common input (e.g., an original image or the output of the previous layer).
%creating high dimensional feature vectors at the output. 
The output of each of these is the aggregation of the outputs of multiple \emph{channels}, where each channel corresponds to a single convolutional filter. Thus, 
in all ConvNet layers, except the last fully connected classification layers, high-dimensional intermediate layer feature vectors can be viewed as a concatenation of 
lower dimensional \textit{subvectors}, each corresponding to the output of a channel.
We make use of this natural ``subvectorization'' in ConvNets
%, where each feature vector is obtained by concatenating the outputs of separate convolution operations, 
to develop a model analysis through channel-wise graph constructions as an alternative to the standard full layer model analysis \cite{huang2018mechanisms, recanatesi2019dimensionality, ansuini2019intrinsic}. This also allows us to estimate model generalization at the channel level. 

In particular, we propose a channel-wise extension of DeepNNK  \cite{shekkizhar2020deepnnk}, CW-DeepNNK, which  leverages the geometrical interpretation and the robustness of NNK \cite{shekkizhar2020} to obtain per channel leave one out~(LOO) classification estimates. 
Our analysis allows us to show that each channel learns specific features, each having different levels of importance for the overall task. Moreover, features extracted by each channel complement each other, so that the combination of all 
subvectors leads to a better classification than that achieved by each individual channel. 
A more detailed study of the properties of channel-wise graphs is carried out in \cite{xBonet21}, where the intrinsic dimension of the data representations at a subvector level and data relationships between channels are studied.

We also conjecture that learning happens at different rates in each channel and  propose a channel-wise early stopping criterion for deep learning systems that does not require any validation data. 
% Note that, though we constrain ourselves to ConvNets, our proposed framework can be easily integrated with any training setups and graph based analysis where the feature vectors involved in graph construction are obtained by the concatenation of several subvectors.
% 
%By using the intermediate data representations at the penultimate layer, 
Training is stopped for a given channel if its corresponding LOO classification performance no longer improves. 
Our proposed method achieves comparable test performance with fewer training iterations than existing methods, such as validation-set-based early stopping and aggregate feature vector based DeepNNK \cite{shekkizhar2020deepnnk}. Our early stopping criterion requires lower runtimes and has the further advantage of not requiring any validation data to be set aside. 
We present strategies to further reduce the complexity of our algorithm while being able to maintain a good estimate of model generalization. 
% equally if not more effective in detecting generalization. 
Our framework is compatible with any optimizer and can be integrated with any ConvNet training setting, while not requiring hyperparameter tuning.
%for a reliable stopping criterion. 
This method will be particularly useful for problems with small datasets, where holding out data for validation is impractical.

\section{Background and Related work}
\subsection{Notation}
We denote scalars, vectors, random variables, and matrices using lowercase (e.g., $x$ and $\theta$), lowercase bold (e.g., $\bS{x}$ and $\bS{\theta}$), uppercase (e.g., $X$ and $Y$), and uppercase bold (e.g., $\bS{K}$ and $\bS{\Phi}$) letters, respectively. 
Wherever applicable, we use a superscript to index and denote a subvector and related measures. For example, a vector $\bS{x}$ in $\mathbb{R}^D$ obtained as the concatenation of $S$ subvectors $\bS{x}_i^s \in \mathbb{R}^{D_s}$ where $\sum_{s=1}^{S} D_s = D$:
\begin{equation}
    \bS{x}_i = 
    \begin{bmatrix}
        \bS{x}_i^1 \\
        \bS{x}_i^2 \\
        \vdots \\
        \bS{x}_i^S \\
    \end{bmatrix} \in \mathbb{R}^D
\end{equation}

$\mathcal{D}_{\text{train}} = \{(\bS{x}_1, y_1), (\bS{x}_2, y_2) \dots (\bS{x}_N, y_N)\}$ is the set of training data and $\mathcal{D}_{\text{train}}^i$ is the set obtained by removing  $(\bS{x}_i, y_i)$ from $\mathcal{D}_{\text{train}}$.
We denote the empirical risk or generalization error associated with a function $\hat{f}$ on $M$ data points $\mathcal{D}$  as
\begin{equation}
    \mathcal{R}_{\text{emp}}(\hat{f} | \mathcal{D}) = \frac{1}{M} \sum_{i} l(\hat{f} (\bS{x}_i), y_i)
\end{equation}
where $\hat{f}(\bS{x})$ is the prediction at $\bS{x}$ and $l(\hat{f} (\bS{x}_i), y_i)$ is the error in the estimate at $\bS{x}_i$ relative to $y_i$.
Given the training data, the leave one out~(LOO) \cite{elisseeff2003leave} estimate of a function at $\bS{x}_i$ is the estimate based on the set containing all training points except $\bS{x}_i$. We denote the risk associated with the LOO procedure as 
\begin{equation}
\label{eq:loo}
    \mathcal{R}_{\text{LOO}}(\hat{f} | \mathcal{D}_{\text{train}}) = \frac{1}{N} \sum_{i=1}^{N} l(\hat{f} (\bS{x}_i) | \mathcal{D}_{\text{train}}^{i}, y_i).
\end{equation}

\subsection{Non Negative Kernel (NNK) regression graphs}
Given $N$ data points represented by feature vectors $\bS{x}$, a graph is constructed by 
%construction method is to obtain an efficient representation of the data by 
connecting each data point (node) to similar data points, so that the weight of an edge  
between two nodes is based on the similarity of the data points, with the absence of an edge (a zero weight) 
denoting least similarity. 
Conventionally, one defines similarity between data points using positive definite kernels \cite{aronszajn1950theory} such as the Gaussian kernel with bandwidth $\sigma$ of (\ref{eq:gaussian}) or range normalized cosine kernel of (\ref{eq:cosine}).
\begin{align} 
    k(\bS{x}_i,\bS{x}_j) &= \text{exp} \left( -\|\bS{x}_i - \bS{x}_j\|^2/2 \sigma^2 \right) \label{eq:gaussian} \\
    k(\bS{x}_i,\bS{x}_j) &= 1/2 + \langle\bS{x}_i,\bS{x}_j\rangle/\left(2 \|\bS{x}_i\| \|\bS{x}_j\| \right) \label{eq:cosine}
\end{align}
Unlike weighted $K$-nearest neighbor (KNN) \cite{dong2011efficient} and $\epsilon$-neighborhood graphs ($\epsilon$-graphs) \cite{chvatal1977aggregations} that are sensitive to the choice of hyperparameters $K/\epsilon$, non negative kernel regression (NNK) graphs \cite{shekkizhar2020} are suggested as a principled approach to graph construction based on a signal representation view. 
% In NNK, graph connectivity at each data (node) is obtained by solving a data approximation problem based on a locally formed dictionary of neighbors. \cite{shekkizhar2020} starts by finding the $K$ nearest neighbors (KNN) of each node denoted by the set $\mathcal{S}$. Then, the NNK optimization at each node solves:
% \begin{equation} \label{eq:nnk}
%  \bS{\theta} = \min_{\bS{\theta}: \; \bS{\theta} \geq 0} \|\phi_i - \bS{\Phi}_\mathcal{S} \bS{\theta} \|^{2}_{_2}
% \end{equation}
% where $\bS{\Phi}_\mathcal{S}$ contains the transformed neighbors and the solution $\bS{\theta}$ corresponds to the NNK weights. 
% Using the \textit{kernel trick}, the NNK problem can be rewritten as: 
% \begin{equation} \label{eq:nnk_problem}
%     \bS{\theta} = \argmin_{\bS{\theta}: \; \bS{\theta} \geq 0} \; \frac{1}{2} \bS{\theta}^T \bS{K}_{\mathcal{S},\mathcal{S}} \bS{\theta} - \bS{K}_{\mathcal{S},i}^T \bS{\theta}
% \end{equation}
% where $\bS{K}_{i,j} = k(\bS{x}_i,\bS{x}_j)$.

\begin{figure*}[!ht]
    \centering
    \includegraphics[width=\textwidth]{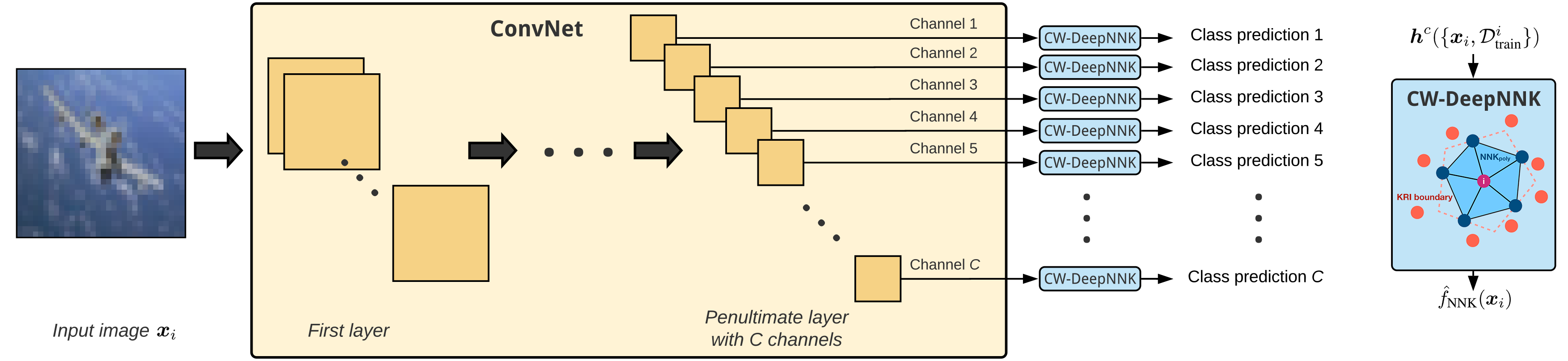}
    \caption{CW-DeepNNK interpolation framework integrated in a ConvNet, replacing the last softmax classification layer. Taking as inputs the activations from individual channels in the penultimate layer, CW-DeepNNK outputs $C$ class predictions for a given input $\bS{x}_i$ via NNK polytope interpolation. Application of the LOO procedure results in $C$ channel-wise generalization estimates which can be used to construct an early stopping criterion without a validation set.}
    \label{fig:cwdeepnnk_scheme}
\end{figure*}

% An advantage of NNK over other methods such as KNN, which select the $K$ largest inner products $\phi^T_i \phi_j$ and can be viewed as a thresholding-based representation, is its robustness to sparsity parameters such as $K$. 
While KNN is still used as an initialization, NNK performs a further optimization akin to orthogonal matching pursuit \cite{tropp2007signal} in kernel space, resulting in a \emph{robust} representation with the added advantage of having a \emph{geometric} interpretation.
The Kernel Ratio Interval (KRI) theorem in \cite{shekkizhar2020} reduces the local NNK graph construction problem  (deciding which of the KNN neighbors of a given node should also be NNK neighbors) into a series of hyper plane conditions, one per NNK weighted neighbor, which applied inductively lead to a convex polytope around each data point, as illustrated in \autoref{fig:cwdeepnnk_scheme} (NNK graph on the right). Intuitively, NNK ignores data that are further away along a \emph{similar} direction as an already chosen point and looks for neighbors in an \emph{orthogonal} direction.
% in the case of the Gaussian kernel \eqref{eq:gaussian}, considering the edge $\theta_{ij}$ connecting node $i$ and node $j$, we can define a hyperplane with normal in the edge direction. The hyperplane divides the space in two, a region $R_{ij}$ that contains $\bS{x}_i$, and its complement $\overline{R}_{ij}$. Then, a third node $k$ will be connected to $i$ only if $\bS{x}_k \in R_{ij}$. 
% % If $\bS{x}_k \in \overline{R}_{ij}$, $\theta_{ik} = 0$.
% %and we say that $k$ has been eliminated by the hyperplane created by $j$.
% The inductive application of the KRI theorem connects to other points while producing a closed decision boundary around $\bS{x}_i$, i.e., the NNK optimization at each node constructs a convex polytope around node $i$ disconnecting all the other points outside the polytope. 
%The resulting set of NNK neighbors for each node is denoted by $\mathcal{N}$. Also note that $\mathcal{N} \subseteq S$.
%: in a three-node scenario, for any positive definite kernel with range $[0,1]$,
%, the necessary and sufficient condition for two data points $\bS{x}_j$ and $\bS{x}_k$ to be connected to $\bS{x}_i$ in an NNK graph is
%\begin{equation} \label{eq:kri}
%    \bS{K}_{j,k} < \frac{\bS{K}_{i,j}}{\bS{K}_{i,k}} < \frac{1}{\bS{K}_{j,k}}
%\end{equation}
%In words, 
%the interval for two candidates, $j$ and $k$, to be connected to the query node $i$ is large if the two candidates are dissimilar. But if the two nodes are very similar, the interval for both edges to exist is very small, and only one node of the two will be connected to the query node $i$.

\subsection{DeepNNK: Neural Networks and NNK interpolation}

DeepNNK \cite{shekkizhar2020deepnnk} is a non-parametric interpolation framework based on local polytopes obtained using NNK graphs \cite{shekkizhar2020} that replaces the standard softmax classification layer of a neural network using the activations obtained at the penultimate layer as input features, for model evaluation and inference. 
% We can continue using the loss obtained with the last layer to perform backpropagation, while the framework can be used to evaluate during training or testing. 
A key advantage of DeepNNK lies in the fact that label interpolation is performed based on the relative positions of points in the training set, which makes it possible to perform leave one out estimation to characterize the task performance of a model and its generalization without the need for additional data, namely the validation set. Note that this would not be possible in the original configuration where classification is performed based on a parametric class boundary.

Formally, the  DeepNNK interpolation estimate at a data point $\bS{x}$ is given by
\begin{equation} \label{eq:nnk_interpolation}
    \hat{f}_{\text{NNK}}(\bS{x}) = \sum_{i \in \text{NNK}_{poly}(\bS{x})} \frac{\theta_i \; y_i}{\sum_{j \in \text{NNK}_{poly}(\bS{x})} \theta_j}
\end{equation}
where $\text{NNK}_{poly}(\bS{x})$ is the convex polytope formed by the NNK identified neighbors and $\bS{\theta}$ corresponds to the NNK weights associated with these neighbors. 
%non-zero recomputed weights obtained from the NNK optimization \eqref{eq:nnk}. 
% Most of the initial $K$ nearest neighbor weights are set to zero and we end up performing the label interpolation with the stable set of NNK neighbors.
Note that in DeepNNK the standard kernels are used to estimate similarity between data points after the non-linear mapping $\bS{h}$, corresponding to a deep neural network~(DNN), has been applied. For example, the Gaussian kernel of \eqref{eq:gaussian} is rewritten as
\begin{equation} \label{eq:gaussian_deep}
    k_{\text{DNN}}(\bS{x}_i,\bS{x}_j) = \text{exp} \left( -\|\bS{h}(\bS{x}_i) - \bS{h}(\bS{x}_j)\|^2 / 2 \sigma^2 \right)
\end{equation}
The authors of \cite{shekkizhar2020deepnnk} show that $\mathcal{R}_{\text{LOO}}(\hat{f}_{\text{NNK}} | \mathcal{D}_{\text{train}})$ can be a better indicator of generalization than the $\mathcal{R}_{\text{emp}}(\hat{f}_{\text{DNN}} | \mathcal{D}_{\text{train}})$ associated with a DNN classifier applied on training data.

% Other methods such as KNN-based interpolation can be biased if the data density is different in different directions in space. In contrast, NNK selects only relevant points for interpolation, eliminating redundant points that do not provide new (\emph{orthogonal}) information.

\subsection{Early stopping methods}

The central idea behind early stopping (stop training or optimization) \cite{prechelt1998early} is that there exists a critical regime during the training of a learning model where the model ceases to generalize (perform better) on unseen data points while being able to do improve performance on given training data. Identifying this point of negative or zero return is also attractive from a computational perspective and is the goal of various early stopping rules or methods in machine learning \cite{yao2007early, raskutti2014early, mahsereci2017early}. 
A conventional and widely popular early stopping method in machine learning is the one based on validation data, which we name as \textit{Validation-based} method. 
Here, one sets aside a part of the training set (referred to as validation set) not to be used for training the model, but on which one only evaluates the performance of the trained model. The validation performance is taken as a proxy for model generalization with training halted when the model begin to perform poorly on the validation set.
% (This was at the beginning of Chapter CW-DeepNNK)
% During the training of a neural network, the goal is to minimize a loss function given a finite training dataset while being able to generalize well on unseen data. To achieve good generalization results, different explicit or implicit regularization methods are used to avoid overfitting the model to the training data \cite{goodfellow2016deep}, including the widely used standard early stopping method, which consists of stopping model optimization based on the model's performance  on a held-out validation set.
% The problem is that there will come a point in the training where the model may stop generalizing and will start to learn the statistical noise of the finite training dataset, which could lead to decreasing  performance on new data, even though the loss function continues to be decreased by additional training \cite{goodfellow2016deep}. But how can we detect when this happens in order to avoid it, and save the model that generalizes better? 
% The most common approach is to hold out part of the training set (a validation set) with which we will not train, but on which we will evaluate the performance of the model so as to obtain a generalization estimate. 
% Then, we can perform early stopping \cite{prechelt1998early}, which consists in stopping the training when we detect that the generalization on the validation set does not improve after some given epochs. Finally, we keep the copy of the model in the iteration where we obtained the best generalization performance.

Although very effective in practice, especially with large training datasets where holding off a small part of the training data has no effect in the learning process, there are drawbacks to Validation-based early stopping \cite{mahsereci2017early}. The validation performance may have a large stochastic error depending on the size of the validation set and may introduce biases leading to poor generalization estimates. A large validation set yields a more robust generalization estimate but may deprive the model of valuable information by reducing significantly the amount of data available for training. 
Thus, in applications with limited training data, it may be better to use all of it for training, rather than holding some of it for validation.
%set is critical and its use would be more valuable if all the data could be used to train the model. 

% Most successful deep learning models include some kind of regularization in their architectures to ensure a small generalization error \cite{zhang2016understanding}. Among them, we find data augmentation, weight decay \cite{krogh1992simple}, dropout \cite{srivastava2014dropout} and batch normalization \cite{ioffe2015batch}. Regularization may also be implicit as in the case of early stopping.

This shortcoming of Validation-based early stopping has led to recent alternatives such as \cite{duvenaud2016early, mahsereci2017early}.
% , most of which are based on a validation set task performance metric, e.g., loss or accuracy curve. The most used criterion is to stop training the model when the observed validation performance has not improved over the best one recorded for a given number of times, commonly called \textit{patience}. Other stopping conditions focus on an absolute change in performance, an average change over a given number of training steps, or the worsening of performance in consecutive steps. An alternative is to stop training when the validation performance is under the best one recorded by a given threshold, while $\mathcal{R}_{\text{emp}}(\mathcal{D}_{\text{train}})$ no longer improves much \cite{prechelt1998early}. But again, these are all different stopping criteria that are generally constructed around the performance curve on an independent validation set, which is the state-of-the-art generalization estimator.
% Other generalization estimates to perform early stopping have been proposed without the need for a validation set. 
% First method
First, \cite{duvenaud2016early} proposed a stopping rule based on estimating the marginal likelihood by tracking the change in entropy of the posterior distribution of the network parameters as an indicator of generalization.
% First method drawbacks
However, the likelihood estimates in this framework are affected when the model has additional regularization terms, which is typically the case for most state-of-the-art methods. 
% Further, the method scales poorly in terms of computational complexity, quickly becoming impractical for large neural network models commonly used today.
% Second method
An alternative approach \cite{mahsereci2017early} presents an early stopping method based on a fast-to-compute local gradient statistic. 
% , aimed at detecting when it represents statistical noise of the finite training set, instead of an informative gradient direction. 
% Second method drawbacks
This method obtains good results compared to the Validation-based method, but requires hyperparameter tuning.
% Both methods drawbacks
Moreover, both \cite{duvenaud2016early} and \cite{mahsereci2017early} rely on gradient-related statistics that are only valid in standard gradient descent or stochastic gradient descent settings and fail to generalize to more advanced optimizers,  such as those based on momentum. Due to their limitations, these methods have been not widely used in practice. A more detailed comparison with our proposed method is carried out in Section \ref{sec:comparison}.

\section{Channel-Wise DeepNNK (CW-DeepNNK)}
\label{sec:channel_deepnnk}

DeepNNK \cite{shekkizhar2020deepnnk} aims at estimating generalization error with the LOO procedure. 
However, the feature vectors used as input are often very high-dimensional, which can lead to poor representation of the ``true'' similarity between data points and therefore to worse performance of polytope-based label interpolation of \eqref{eq:nnk_interpolation}. Partially due to this,  estimation of generalization performance (on an unseen dataset) using \eqref{eq:nnk_interpolation} does not obtain achieve the same accuracy as the conventional procedure based on using a separate validation set.

To address this problem, 
in this work we propose a local polytope label interpolation in individual channels (CW-DeepNNK), as illustrated in \autoref{fig:cwdeepnnk_scheme}. Instead of using the transformed data representations of the full penultimate layer $\bS{h}(\bS{x})$, which consists of the aggregation of outputs of $C$ convolutional channels, we propose dividing the feature space into channels:
\begin{equation}
    \bS{h}(\bS{x}_i) = 
    \begin{bmatrix}
        \bS{h}^1(\bS{x}_i) \\
        \bS{h}^2(\bS{x}_i) \\
        \vdots \\
        \bS{h}^C(\bS{x}_i) \\
    \end{bmatrix} \in \mathbb{R}^{D_{\bS{h}}},
\end{equation}
which are well-defined and can be interpreted individually. Then, in each channel $c$, the first step for the CW-DeepNNK LOO procedure is to use the intermediate representations $\bS{h}^{c}(\mathcal{D}_{\text{train}}^i)$ as feature vectors to construct an NNK neighborhood for each data point $\bS{h}^{c}(\bS{x}_i)$ in the training set.
%to perform the KNN search, obtaining $\mathcal{S}^{c}$
%The second step is to construct the similarity matrix $\bS{K}_{\mathcal{S}^{c}}$, and solve the NNK regression \eqref{eq:nnk_problem}. 
Second, perform the NNK interpolation \eqref{eq:nnk_interpolation}. Finally, we compute the LOO estimation \eqref{eq:loo} per channel, obtaining the CW-DeepNNK label interpolation errors $\mathcal{R}_{\text{LOO}}^1, \mathcal{R}_{\text{LOO}}^2, \dots, \mathcal{R}_{\text{LOO}}^C$.

By using the CW-DeepNNK procedure at each training epoch we obtain a label interpolation error curve for each channel, which allow us to monitor the generalization of the model during training. Then, we propose a novel channel-wise early stopping criterion described in Algorithm \ref{alg:cwdeepnnk_stopping}, which does not require a validation set and the stopping is performed in stages, allowing us to stop the training of each channel independently. Note that this is the first early stopping criterion that (i) does not require a separate validation set and (ii) is based on a task performance metric (e.g., accuracy). 
In contrast, the other proposed early stopping methods without a validation set \cite{mahsereci2017early, duvenaud2016early} based their generalization estimation on properties of the gradients computed during training.
%, instead of estimates of task performance.

\begin{algorithm}[htbp]
\caption{CW-DeepNNK progressive early stopping without a validation set}\label{alg:cwdeepnnk_stopping}

\textbf{Input:} 

\hspace*{\algorithmicindent} $\mathcal{D}_{\text{train}} = \{(\bS{x}_1, y_1), (\bS{x}_2, y_2) \dots (\bS{x}_N, y_N)\}$: training set

\hspace*{\algorithmicindent} $\bS{w}$: model parameters \Comment{$\bS{w}_\text{penult}$: penultimate layer params}

\hspace*{\algorithmicindent} $\bS{h}$: model non-linear mapping at the penultimate layer

\hspace*{\algorithmicindent} $C$: number of channels in penultimate layer

\hspace*{\algorithmicindent} $n$: number of steps between generalization evaluations

\hspace*{\algorithmicindent} $p$: \textit{patience}, number of times to observe worsening

\hspace*{\algorithmicindent} LOO NNK interpolation error before stopping channel

\hspace*{\algorithmicindent} $K$: number of initial neighbors

\textbf{Output:} best parameters $\bS{w}^*$, best number of training steps $t^*$

\begin{algorithmic}[1]
\State $t = 0$, 
$\bS{q} = \bS{p}$, $\bS{r} = \bS{\infty}$
\State \textbf{initialize} $\bS{w}$
\State $\bS{w}^* = \bS{w}$, $t^* = t$
\While{$\bS{q} \neq \bS{0}$}
\State Update $\bS{w}$ by running the training algorithm for $n$ steps
\State $t = t+n$
\For{$c=1:C$}
\If{$\bS{q}(c) > 0$}
\For{$i=1:N$}
\State $\bS{\theta} = \textit{\footnotesize findNNKneighbors}(\bS{h}^{c}(\{\bS{x}_i, \mathcal{D}_{\text{train}}^i\}), K)$
\State $\hat{f}_{\text{NNK}}(\bS{x}_i) = \sum_{j} \frac{\theta_j y_j} {\sum_{l} \theta_l} $
\EndFor
\State $ \mathcal{R}_{\text{LOO}}^{c} = \frac{1}{N} \sum_{i} l(\hat{f}_{\text{NNK}} (\bS{x}_i) | \mathcal{D}_{\text{train}}^{i}, y_i )$
\If{$\mathcal{R}_{\text{LOO}}^{c} < \bS{r}(c)$}
\State $\bS{r}(c) = \mathcal{R}_{\text{LOO}}^{c}$, $\bS{q}(c) = p$
\State $\bS{w}^* = \bS{w}$, $t^* = t$
\Else
\State $\bS{q}(c) = \bS{q}(c) - 1$
\EndIf
\If{$\bS{q}(c) = 0$}
\State Freeze and stop training $\bS{w}_{\text{penult}}^{c}$
\EndIf
\EndIf
\EndFor
\EndWhile
\end{algorithmic}
\end{algorithm}

\subsection{Channel-wise early stopping without a validation set}
\label{sec:early_stopping_alg}

Starting from the standard \textit{patience} criterion, we monitor the generalization performance in the penultimate layer channels and we use a patience parameter $p$ in each channel. When a channel stops generalizing, i.e., $\mathcal{R}_{\text{LOO}}^{c}$ has not improved in $p$ observations, we freeze the model parameters of the channel and stop training it. 
The rest of the model continues learning until each of the channels stops generalizing, where we consider that we have reached the optimal point and the overall generalization of the model no longer improves. 
Finally, we save the best model parameters $\bS{w}^*$ where the last minimum generalization error is detected. 
Code for the proposed method is available 
online.\footnote{ \url{https://github.com/STAC-USC/CW-DeepNNK_Early_Stopping}}

\subsection{Complexity}
The channel-wise LOO label interpolation \textit{baseline} is computationally expensive, since it requires constructing one NNK graph per training instance, per channel, and per epoch. However, several improvements can be made to achieve an overall lower computation cost, competitive with the state-of-the-art early stopping methods in large scale problems.
First, performing NNK independently in every epoch is very costly, but the features learned in a given channel in consecutive epochs are similar. Thus, complexity can be reduced by a factor of $T$ by performing one LOO estimation every $T$ epochs. As an alternative, we can maintain a fixed NNK assignment for all nodes and monitor the approximation error for node $i$ as a function of its neighbors. When this error increases a new NNK graph is constructed, so that a better set of neighbors for node $i$ can be found. 

Other improvements in efficiency can be achieved by reducing the number of channels and the number of nodes for which LOO label interpolation is performed.
For example, we could perform the interpolation only on those channels with lowest error, as described in Section \ref{sec:relevant_channels}. 
We could also perform random subsampling of training data points instead of performing the full LOO procedure. 
As a result, computing time would be drastically reduced, at the expense of less reliable and more random generalization 
estimates, 
whose impact could be reduced by augmenting the patience parameter, which would make the actual stopping point less sensitive to random oscillations in the estimates.
%thus avoiding curve oscillations and finding the global generalization error minimum with higher probability.

Additionally, while selecting $K$ for the experiments we observed that both NNK-based methods are very robust with respect to the selection of this single hyperparameter. Therefore, $K$ could be chosen as the minimum $K$ that yields stable results, significantly reducing the complexity. Exploring all these ideas for further efficiency improvement of the proposed method is left for future work.

\section{Experiments}
\label{sec:exp_2}

\begin{figure*}[!ht]
    \centering
    \includegraphics[width=\textwidth]{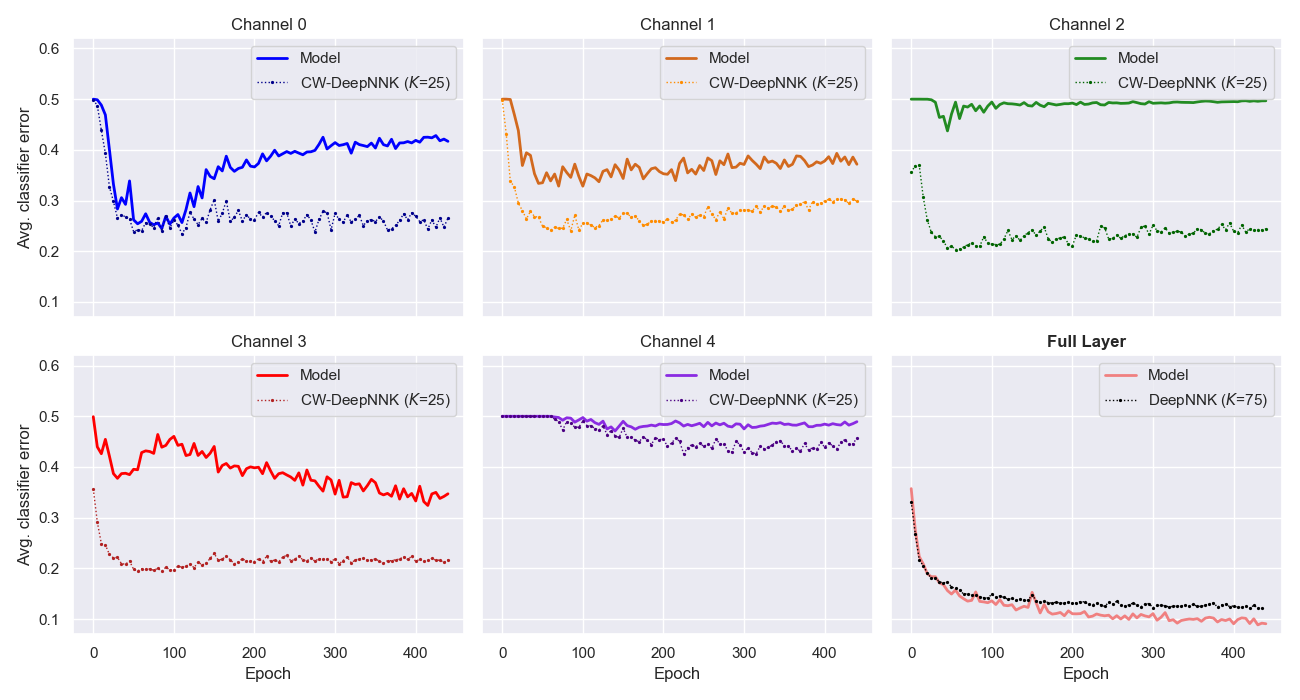}
    \caption{Classification error using interpolating classifiers (DeepNNK, CW-DeepNNK) and model error on training data. CW-DeepNNK significantly outperforms the softmax classification of the neural network with only a single channel of the penultimate layer  activated. Further, most channels attain their CW-DeepNNK minima at an early number of iterations with performance deteriorating later indicative of overfitting.}
    \label{fig:channel_deepnnk_vs_model}
\end{figure*}

% TODO: State overview and connect it to model description

In this section, we evaluate CW-DeepNNK for estimating channel generalization and construct an efficient early stopping method with several benefits over DeepNNK and the Validation-based method. We focus our experiments on a binary classification setting using 2 classes of CIFAR-10 dataset \cite{krizhevsky2009learning}: ``plane'' and ``ship''. Training set and validation set, if needed, are split in a stratified fashion in each experiment ensuring class balance in each set.
We consider a 7 layer ConvNet architecture consisting of 4 convolution layers with 5 depth channels and ReLU activations, 2 max-pool layers and a fully connected softmax layer. We train the model with the Adam optimizer \cite{kingma2015adam}, with a learning rate of 0.001 and batch size of 50 on a GTX Titan X with 8GB of memory. 
We compare the model performance (based on softmax classification) with that of DeepNNK and CW-DeepNNK interpolation on train data. We observe that some channels learn more valuable features than others for the classification task (see \autoref{fig:channel_deepnnk_vs_model}). We analyze the behaviour of our generalization estimate in the presence and absence of dropout regularization in Section \ref{sec:regul_vs_nonregul} and show how one can identify channel importance based on NNK's local polytope geometry and activation patterns in Section \ref{sec:relevant_channels}.
Finally, in Section \ref{sec:comparison} we compare NNK-based generalization estimates with that of the Validation-based strategy \cite{prechelt1998early} to perform early stopping and discuss the complexity of each.

% explain how I will use the network in the experiments, comparing regularized (using dropout with a rate of 0.2) and non-regularized behaviours, and comparing model performance on train data with the LOO DeepNNK interpolation performance, where we replace the last layer of the model tup, we replace the last classification layer with this interpolator framework, using the full output of the last convolutional layer or individual channels.

\subsection{Generalization estimates using CW-DeepNNK}
\label{sec:gen_estimates}

%A convolutional layer is composed of various channels that are outputs of different filtering operations. 
Each channel of a convolutional layer defines a feature subspace where we should be able to quantify how useful the information from that channel is for the classification task, and have a better interpretation of the captured features in each channel. 
%We can assume that each filter captures different features, although they are not completely independent between channels.

\autoref{fig:channel_deepnnk_vs_model} shows a comparison between model error on training data, DeepNNK and CW-DeepNNK label interpolation error with LOO estimation. In the DeepNNK case, the error gap between the model on train data and LOO DeepNNK increases with the epochs, indicating that the generalization performance is worsening and the model starts to overfit to the train data. Also, we can see how CW-DeepNNK has a much better performance than the model when only a single channel is activated. Note that an error of 0.5 in a binary classification is as bad as doing random classification. 
We also observe how the interpolation error in the channels soon reaches a minimum, and then the classification error increases again. This minimum may indicate the optimal point of generalization in each channel, from which the learned features begin to overfit the training data.

Although the last fully connected layer of the model is trained to use the full combination of features from all channels, we wanted to see what happens if the model has to perform classification when relying only on partial information. \autoref{fig:channel_deepnnk_vs_model} shows how our method is able to perform much better in each independent channel subspace, and the model is not capable of performing at a decent level when some feature channels are deactivated. Thus, we now have a new point of view, where we can estimate properly how useful each of the individual channels is for the task.

\subsection{Regularized vs. Non-Regularized models} \label{sec:regul_vs_nonregul}

\begin{figure*}[ht] 
    \centering
    \begin{subfigure}[b]{0.495\textwidth}  
        \centering 
        \includegraphics[width=\textwidth]{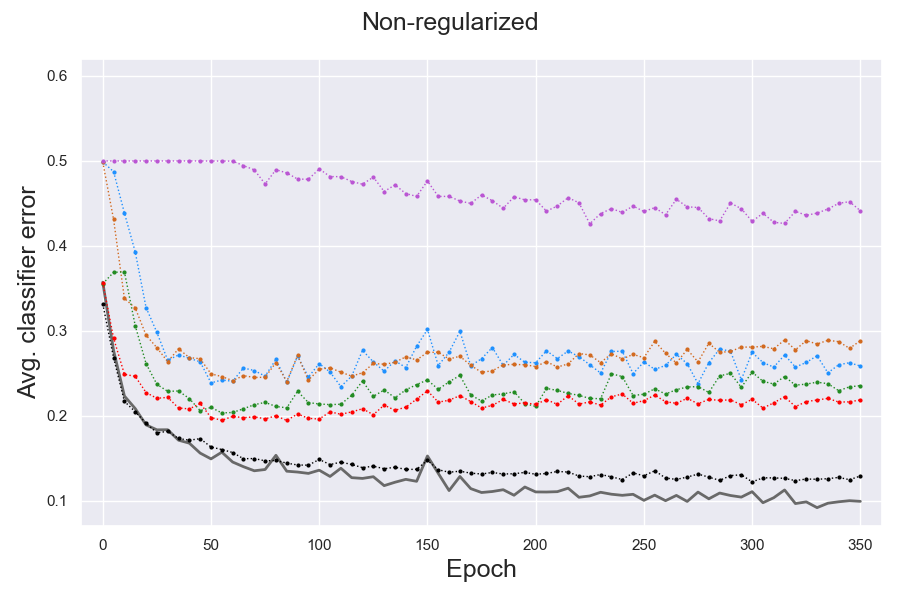}
        %\caption{ }    
        \label{fig:LOO_nonregul}
    \end{subfigure}
    \hfill
    \begin{subfigure}[b]{0.495\textwidth}
        \centering
        \includegraphics[width=\textwidth]{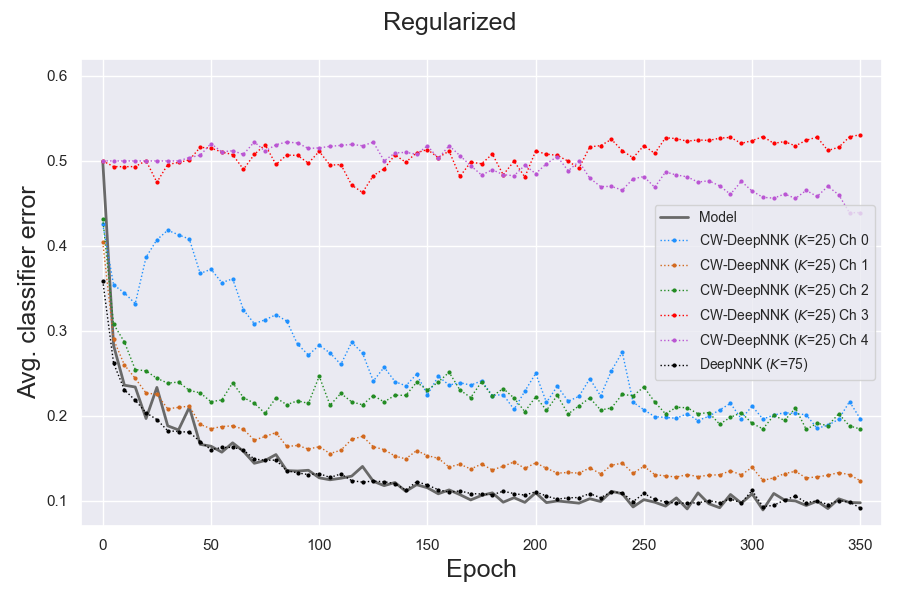}
        %\caption{ }
        \label{fig:LOO_regul}
    \end{subfigure}
    \vskip\baselineskip
    \vspace{-10mm}
    \begin{subfigure}[b]{0.495\textwidth}   
        \centering 
        \includegraphics[width=\textwidth]{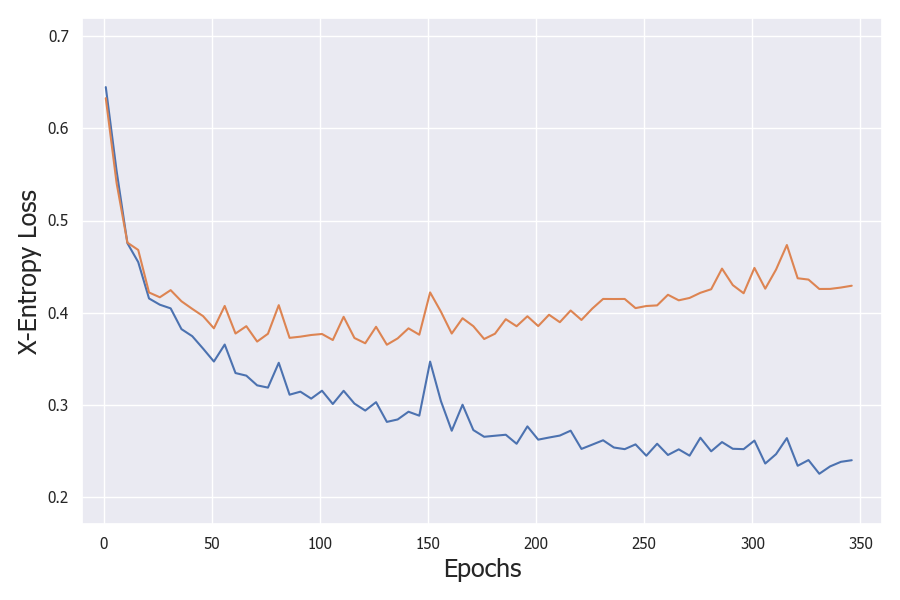}
        %\caption{ }
        \label{fig:model_loss_nonregul}
    \end{subfigure}
    \hfill
    \begin{subfigure}[b]{0.495\textwidth}   
        \centering 
        \includegraphics[width=\textwidth]{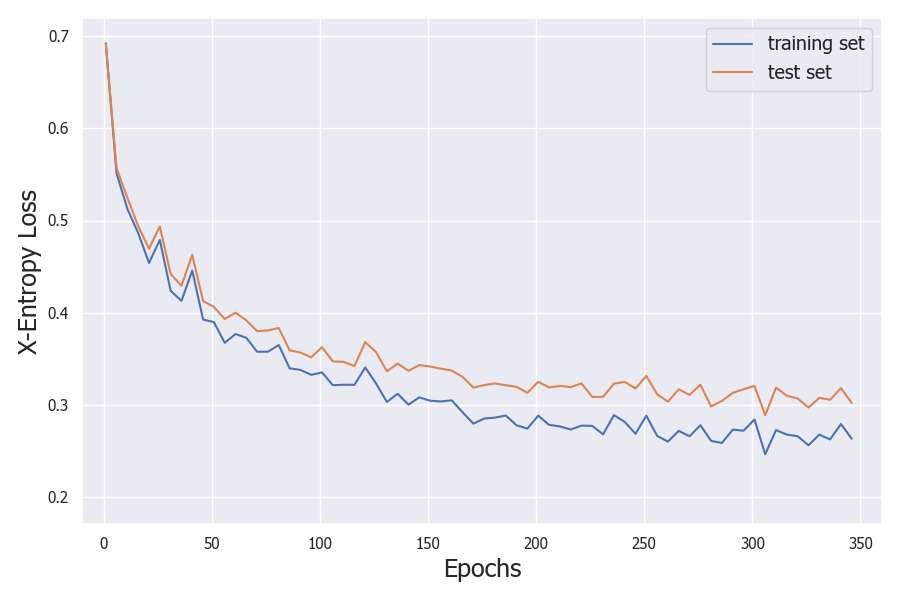}
        %\caption{ }  
        \label{fig:model_loss_regul}
    \end{subfigure}
    \vspace{-10mm}
    \captionsetup{justification=centering}
    \caption{\textbf{top row}: DeepNNK, CW-DeepNNK and training model error; \textbf{bottom row}: model loss on train and test data, metrics shown every 5 epochs. \textbf{left column}: no regularization; \textbf{right column}: dropout with a rate of 0.2 in each layer. With no regularization, most CW-DeepNNK minima coincide with that of the test loss, where early stopping should occur. In a well-regularized model, CW-DeepNNK continues improving as does the model in the test set.}
    \label{fig:LOO_regul_comparison}
\end{figure*}

% Introduce experiment motivation
We can analyze the performance of CW-DeepNNK in different scenarios and compare the  results  in Section \ref{sec:gen_estimates}, with no regularization, to results obtained using explicit regularization, adding a dropout layer \cite{srivastava2014dropout} after each convolutional layer, with a dropout rate of 0.2.
In both cases we see individual channels that fail to learn features of the data relevant to the task, obtaining almost maximum error in binary classification $\mathcal{R}_{\text{LOO}} \approx 0.5$. In the following study we focus on the channels with an error $\mathcal{R}_{\text{LOO}} < 0.4$.

\autoref{fig:LOO_regul_comparison} shows how in a non-regularized model, our CW-DeepNNK generalization error estimate finds a minimum at an early number of iterations. If we compare it with the test performance, we see that channel-wise LOO performances peak at a similar place to the peak of the test loss, where early stopping would occur using the Validation-based method. 
In addition, channel minimum error detection in different points in time suggests that early stopping could be performed progressively by channels.
Further, monitoring this generalization estimator in a well-regularized network that does not overfit, our estimator is consistent, observing a performance curve similar to that of the test set.
%All this indicates that this metric is a good estimator of generalization and could replace other estimators such as validation accuracy for performing early stopping. 

% Reduce text and say it is reliable under various conditions.

We can also detect that in the non-regularized case, the \textit{important} channels, i.e., those channels where best CW-DeepNNK performances are achieved, have a very similar but very poor performance with an error between 0.2 and 0.3. Instead, in the regularized case we can see that the important channels reach an error below 0.2.

\subsection{Channel importance estimation using NNK polytopes} \label{sec:relevant_channels}

As seen in Section \ref{sec:regul_vs_nonregul}, only some channels of a convolutional layer concentrate the data features that are key for  the task, and we can detect these channels before performing the LOO interpolation based on the NNK polytope local geometry and zero patterns in the activations.

\begin{figure*}[htb]
    \centering % <-- added
\begin{subfigure}{0.32\textwidth}
  \includegraphics[width=\linewidth]{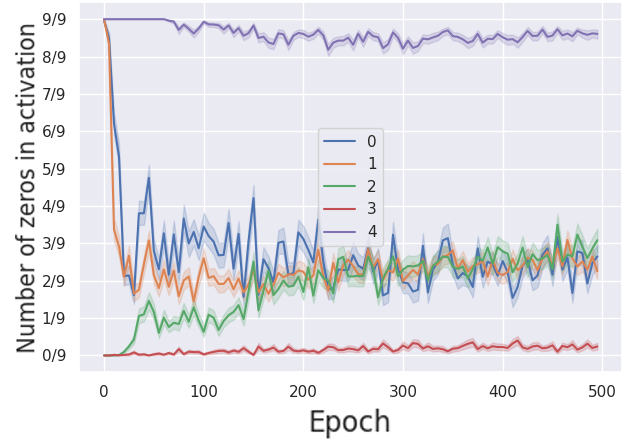}
  %\caption{Number of zeros per channel, \\non-regularized case.}
  \label{fig:zeros_nonregul}
\end{subfigure}\hfil % <-- added
\begin{subfigure}{0.32\textwidth}
  \includegraphics[width=\linewidth]{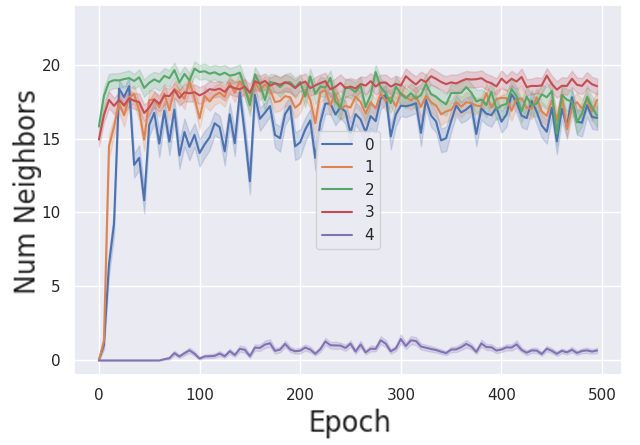}
  %\caption{Number of NNK neighbors per channel, non-regularized case.}
  \label{fig:numNNK_nonregul}
\end{subfigure}\hfil % <-- added
\begin{subfigure}{0.32\textwidth}
  \includegraphics[width=\linewidth]{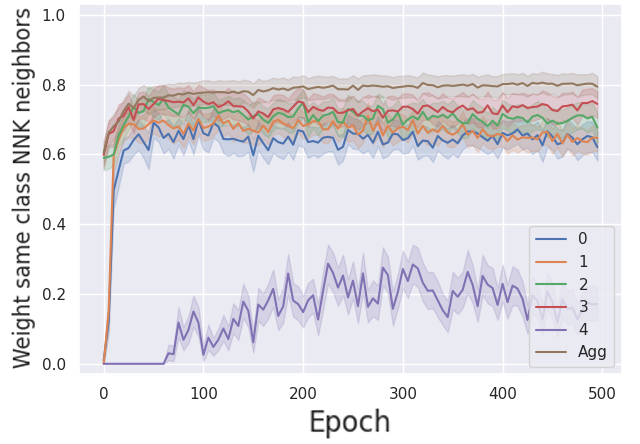}
  %\caption{Weight same class NNK neighbors, non-regularized case.}
  \label{fig:weightNNK_nonregul}
\end{subfigure}
\medskip
\vspace{-5mm}
\begin{subfigure}{0.32\textwidth}
  \includegraphics[width=\linewidth]{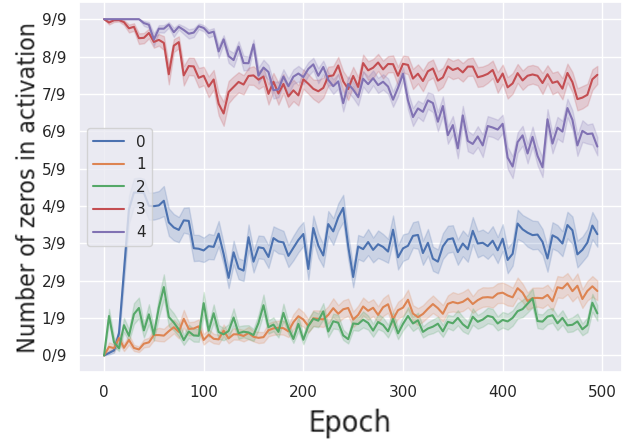}
  %\caption{Num. of zeros per channel, regularized case.}
  \label{fig:zeros_regul}
\end{subfigure}\hfil % <-- added
\begin{subfigure}{0.32\textwidth}
  \includegraphics[width=\linewidth]{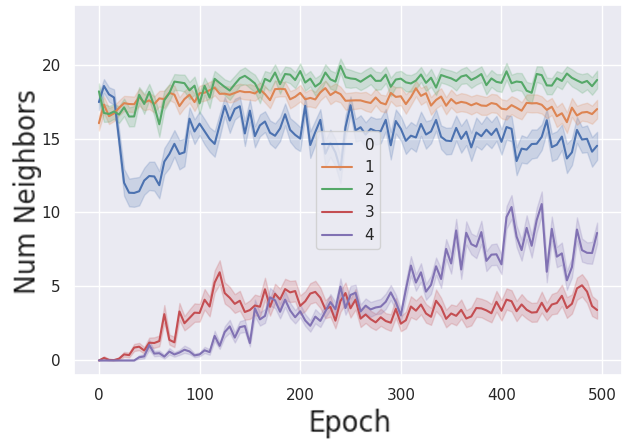}
  %\caption{Num. of NNK neighbors per channel, regularized case.}
  \label{fig:numNNK_regul}
\end{subfigure}\hfil % <-- added
\begin{subfigure}{0.32\textwidth}
  \includegraphics[width=\linewidth]{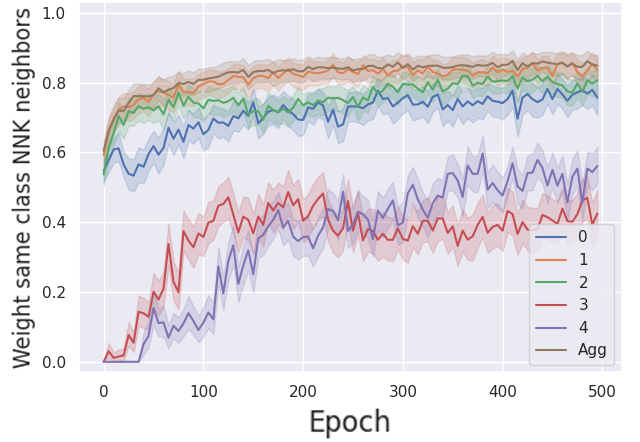}
  %\caption{Weight same class NNK neighbors, regularized case.}
  \label{fig:weightNNK_regul}
\end{subfigure}
\vspace{-5mm}
\captionsetup{justification=centering}
\caption{\textbf{left column}: number of zeros per channel; \textbf{middle column}: number of NNK neighbors per channel; \textbf{right column}: total weight of same class NNK neighbors.
\textbf{top row}: non-regularized case; \textbf{bottom row}: regularized case. Important channels for the task generally present a low number of zeros in their activations, also obtaining larger NNK graphs with a high weight of same-class neighbors to the query node.}
\label{fig:relevant_channel_detection}
\end{figure*}

In \autoref{fig:relevant_channel_detection} we can see how in the non-regularized case (first row), there is a very noticeable binary behavior. 4 out of 5 channels present very few zeros in their activations due to ReLU, i.e. in most dimensions we will find positive values. In these same channels, nodes in the NNK graph tend to be connected to a higher number of neighbors, obtaining a high weight of NNK neighbors of the same class to the instance. 
In the remaining channel, we observe that most dimensions are always zero so the ID is very low, nodes tend to be connected to a lower number of NNK neighbors and the weight of neighbors of the same class is very low, since the information in that channel is very poor. 
These properties will lead to a bad label interpolation. 
In the regularized case (second row) we observe an heterogeneous behavior: there are channels with less zero dimensions than others, in which we will connect with more NNK neighbors and we will obtain a higher weight of neighbors of the same class.
What is relevant in this case is that, in the channels with higher NNK dimensionality we obtain a higher weight of neighbors of the same class, surpassing the weight obtained in the non-regularized case, thus leading to lower NNK label interpolation error, which is indicative of a better generalization in the regularized case.

In short, we can predict the channel LOO label interpolation performance of the network (\autoref{fig:LOO_regul_comparison}) by analyzing the dimensionality and NNK graph local geometry in each channel. 
In channels with fewer zero dimensions we will obtain a higher dimensional NNK graph with higher same-class weights, which will lead to better interpolation. 
Besides, the non-regularized networks have a more homogeneous behavior between channels but with worse performance in general, while for  regularized models, the most important channels drive a better overall result.

\subsection{Early stopping with CW-DeepNNK}
\label{sec:comparison}

We compare our generalization estimate with the Validation-based method as well as with the DeepNNK estimate \cite{shekkizhar2020deepnnk}, to perform early stopping. 
In this case, we use a patience parameter as stopping rule for the Validation-based method and for the full layer LOO DeepNNK interpolation. 
For our estimate, we use a patience stopping rule in each channel, as described in Section \ref{sec:early_stopping_alg}. Other non-validation set methods based on gradient-related statistics \cite{mahsereci2017early, duvenaud2016early} are promising approaches, but unlike our proposed method based on a task performance metric, they are not compatible with momentum-based optimizers and require hyperparameter tuning for a reliable stopping. Moreover, \cite{mahsereci2017early} assumes that all weights of a layer will converge at similar speeds, which may not be necessarily true in convolutional layers with multiple channels. The criterion proposed in \cite{duvenaud2016early} 
%is the only method with available code, and 
has only been validated on simpler models, e.g., one hidden layer networks. 
%Also, we were not successful in applying it to our problem, and further refinements are needed to build a practical stopping rule with the marginal likelihood estimator.

Note that other stopping rules \cite{prechelt1998early, terry2021statistically} could use our generalization estimate as a substitute for the validation curve, but we leave these experiments for future work. 

Even though weights of the penultimate layer are prevented from further training in our proposed channel-wise method, this does not result in extra benefits in computing complexity as the fraction of parameters that we stop training in intermediate stages is not significant. An extension of this method to the rest of the layers of the model could speed up the training iterations. 

\begin{figure*}[ht]
    \centering
    \begin{subfigure}[b]{\textwidth}  
        \centering 
        \includegraphics[width=\textwidth]{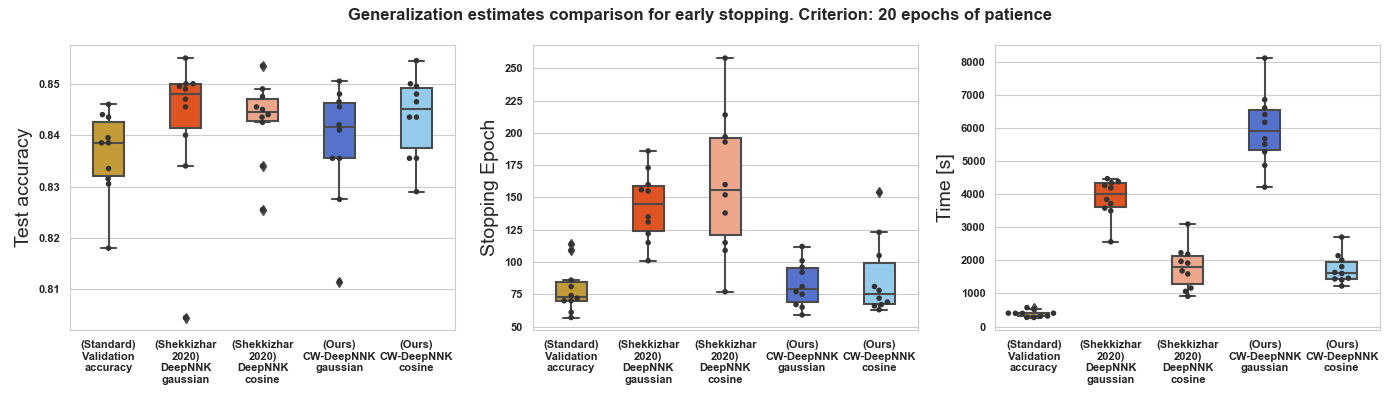}
   
        \caption{Validation-based method using 20\% of labeled samples as validation set, and DeepNNK and CW-DeepNNK for both Gaussian kernel \eqref{eq:gaussian} and cosine kernel \eqref{eq:cosine}. NNK-based methods obtain higher test accuracies than the Validation-based method, training the model with all labeled data. CW-DeepNNK cosine obtains the best trade-off between test accuracy and training iterations with an overhead of computational time.}
        \label{fig:early_stopping_comparison}
    \end{subfigure}
    \begin{subfigure}[b]{\textwidth}  
        \centering
        \includegraphics[width=\textwidth]{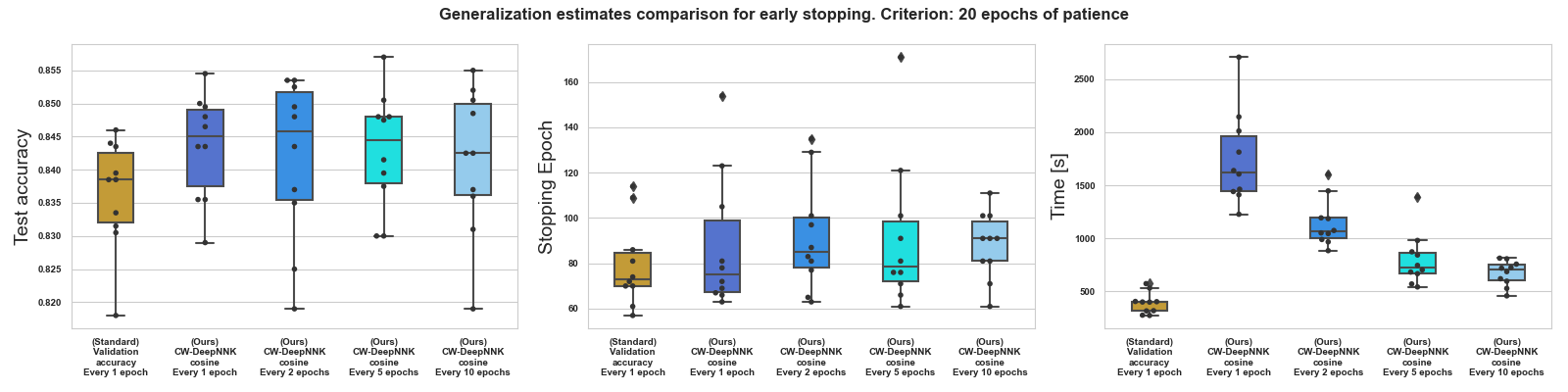}
   
        \caption{Comparison of LOO computation frequencies for CW-DeepNNK; Validation-based method using 20\% of labeled samples as validation set. CW-DeepNNK frequency reduction preserves superior results while requiring less computation time, comparable to the Validation-based method.}
        \label{fig:early_stopping_reduce_freq}
    \end{subfigure}
    \begin{subfigure}[b]{\textwidth}  
        \centering
        \includegraphics[width=\textwidth]{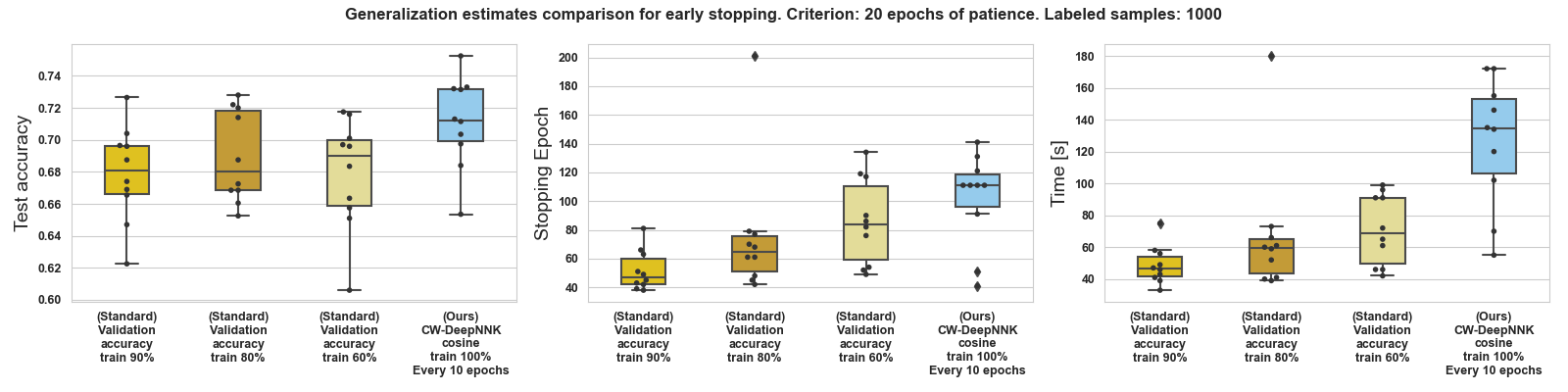}
   
        \caption{Small dataset case with 1000 labeled samples, obtained from random sampling at each initialization. CW-DeepNNK obtains higher accuracies than the Validation-based method, using all labeled data available for training.}
        \label{fig:early_stopping_mini_dataset}
    \end{subfigure}
    \captionsetup{justification=centering}
    \caption{Test accuracy, stopping epoch and training elapsed time for different early stopping methods. Note that test accuracy refers to the accuracy obtained in the test set with the best model according to each criterion, i.e., where we find the last minimum generalization error, not the last version where we stop training. $K=15$ for NNK-based methods.}
    \label{}
\end{figure*}
  
\autoref{fig:early_stopping_comparison} shows the results obtained using the different generalization estimates for early stopping, with a patience of 20 epochs and 10 different model initializations and validation set partitions. We can see how all NNK-based methods obtain test accuracies higher than the Validation-based method. Although the best models are obtained in the DeepNNK case, it requires a lot of epochs to find the optimal stopping point whereas the other methods are able to detect overfitting in a much earlier stage. In the two proposed methods based on NNK interpolation, using a patience parameter that is too small can lead to not finding the global minimum of generalization error, leading to premature stopping and lower test accuracy, as in the case of the outliers. Therefore, choosing a higher patience can ensure a high test accuracy, but at the cost of a longer training, often unnecessary. 
The proposed channel-wise generalization estimate using the range normalized cosine kernel \eqref{eq:cosine} is the alternative that obtains the best trade-off between performance, training iterations and time.
% The NNK-based approaches entail a higher complexity using the Gaussian kernel \eqref{eq:gaussian}. However, using the range normalized cosine kernel \eqref{eq:cosine} we reach computation times comparable to the standard method. 
In \autoref{fig:early_stopping_reduce_freq} we study different LOO computation frequencies to further improve the efficiency of the proposed algorithm, maintaining a total wait of 20 epochs in all cases. Test accuracy and stopping epochs are preserved while the computation time is significantly reduced with increasing $T$, reaching an efficiency and results similar to those of the Validation-based method in the case of $T=10$.

We also study the case of dealing with small datasets, where using the available data wisely is critical for good generalization and test performance. Using a labeled data subset of 1000 samples of the two CIFAR-10 classes (split between train set and validation set if required) in \autoref{fig:early_stopping_mini_dataset} we see how our method obtains state-of-the-art results outperforming the standard Validation-based approach, since we can train the model with all labeled data without the need of separating data for a validation set.

\section{Conclusion and Future work}

% 1. Generalization estimate
We introduced a novel approach for channel-wise generalization estimation in ConvNets based on local NNK polytope interpolation. From an in-depth analysis of the polytope local geometry and data representation patterns, we showed how to detect the channels that are the most important for the task and obtain best interpolation performance.
% 2. Early stopping
We also presented a progressive channel-based early stopping strategy which does not require a validation set. Our approach may be the preferred for early stopping in situations where test performance is key or when labeled data is scarce, since we 
would not need to hold out data for a validation set. Based on a task performance metric, as is the widely used Validation-based method, the presented criterion can be integrated into any training setting (including momentum-based optimizers and explicit regularization methods) and
does not require hyperparameter tuning.
% it leads to better interpretability of the generalization estimate than the other no-validation set methods based on gradient-related statistics.

% Future work

Future work should explore the ideas presented to further improve efficiency, obtaining competitive results on large-scale problems. Future research could also be in the direction of neural network pruning based on the CW-DeepNNK interpolation error, since we have observed that in certain channels we have practically no useful information to perform the interpolation, and those channels could be pruned, resulting in a more compact network with less computational cost. Another line of research could involve a \textit{progressive} early stopping of the full model, achieving a significant saving of gradient computation and backpropagation throughout the training.

% \typeout{} 
\bibliographystyle{ieeetr}
\bibliography{root.bib}

\begin{thebibliography}{10}

\bibitem{ortega2018graph}
A.~Ortega, P.~Frossard, J.~Kova{\v{c}}evi{\'c}, J.~M. Moura, and
  P.~Vandergheynst, ``Graph signal processing: Overview, challenges, and
  applications,'' {\em Proceedings of the IEEE}, vol.~106, no.~5, pp.~808--828,
  2018.

\bibitem{ortega_2021}
A.~Ortega, {\em Introduction to Graph Signal Processing}.
\newblock Cambridge University Press, 2021.

\bibitem{bacciu2020gentle}
D.~Bacciu, F.~Errica, A.~Micheli, and M.~Podda, ``A gentle introduction to deep
  learning for graphs,'' {\em Neural Networks}, 2020.

\bibitem{lassance2021laplacian}
C.~Lassance, V.~Gripon, and A.~Ortega, ``Laplacian networks: Bounding indicator
  function smoothness for neural networks robustness,'' {\em APSIPA
  Transactions on Signal and Information Processing}, vol.~10, 2021.

\bibitem{lassance2021representing}
C.~Lassance, V.~Gripon, and A.~Ortega, ``Representing deep neural networks
  latent space geometries with graphs,'' {\em Algorithms}, vol.~14, no.~2,
  p.~39, 2021.

\bibitem{lassance2020deep}
C.~Lassance, M.~Bontonou, G.~B. Hacene, V.~Gripon, J.~Tang, and A.~Ortega,
  ``Deep geometric knowledge distillation with graphs,'' in {\em ICASSP
  2020-2020 IEEE International Conference on Acoustics, Speech and Signal
  Processing (ICASSP)}, pp.~8484--8488, IEEE, 2020.

\bibitem{shekkizhar2020deepnnk}
S.~Shekkizhar and A.~Ortega, ``Deepnnk: Explaining deep models and their
  generalization using polytope interpolation,'' {\em arXiv preprint
  arXiv:2007.10505}, 2020.

\bibitem{gripon2018inside}
V.~Gripon, A.~Ortega, and B.~Girault, ``An inside look at deep neural networks
  using graph signal processing,'' in {\em 2018 Information Theory and
  Applications Workshop (ITA)}, pp.~1--9, IEEE, 2018.

\bibitem{papernot2018}
N.~Papernot and P.~McDaniel, ``Deep k-nearest neighbors: Towards confident,
  interpretable and robust deep learning,'' {\em arXiv preprint
  arXiv:1803.04765}, 2018.

\bibitem{shekkizhar2020}
S.~Shekkizhar and A.~Ortega, ``Graph construction from data by non-negative
  kernel regression,'' in {\em 2020 IEEE Int. Conf. on Acoustics, Speech and
  Signal Process. (ICASSP)}, pp.~3892--3896, IEEE, 2020.

\bibitem{he2016deep}
K.~He, X.~Zhang, S.~Ren, and J.~Sun, ``Deep residual learning for image
  recognition,'' in {\em Proceedings of the IEEE conference on computer vision
  and pattern recognition}, pp.~770--778, 2016.

\bibitem{zagoruyko2016wide}
S.~Zagoruyko and N.~Komodakis, ``Wide residual networks,'' in {\em British
  Machine Vision Conference 2016}, British Machine Vision Association, 2016.

\bibitem{tan2019efficientnet}
M.~Tan and Q.~Le, ``Efficientnet: Rethinking model scaling for convolutional
  neural networks,'' in {\em International Conference on Machine Learning},
  pp.~6105--6114, PMLR, 2019.

\bibitem{huang2018mechanisms}
H.~Huang, ``Mechanisms of dimensionality reduction and decorrelation in deep
  neural networks,'' {\em Physical Review E}, vol.~98, no.~6, p.~062313, 2018.

\bibitem{recanatesi2019dimensionality}
S.~Recanatesi, M.~Farrell, M.~Advani, T.~Moore, G.~Lajoie, and E.~Shea-Brown,
  ``Dimensionality compression and expansion in deep neural networks,'' {\em
  arXiv preprint arXiv:1906.00443}, 2019.

\bibitem{ansuini2019intrinsic}
A.~Ansuini, A.~Laio, J.~H. Macke, and D.~Zoccolan, ``Intrinsic dimension of
  data representations in deep neural networks,'' {\em Advances in Neural
  Information Processing Systems}, vol.~32, pp.~6111--6122, 2019.

\bibitem{xBonet21}
D.~Bonet, ``Improved neural network generalization using channel-wise nnk graph
  constructions,'' 2021.

\bibitem{elisseeff2003leave}
A.~Elisseeff, M.~Pontil, {\em et~al.}, ``Leave-one-out error and stability of
  learning algorithms with applications,'' {\em NATO science series sub series
  iii computer and systems sciences}, vol.~190, pp.~111--130, 2003.

\bibitem{aronszajn1950theory}
N.~Aronszajn, ``Theory of reproducing kernels,'' {\em Transactions of the
  American mathematical society}, vol.~68, no.~3, pp.~337--404, 1950.

\bibitem{dong2011efficient}
W.~Dong, C.~Moses, and K.~Li, ``Efficient k-nearest neighbor graph construction
  for generic similarity measures,'' in {\em Proceedings of the 20th
  international conference on World wide web}, pp.~577--586, 2011.

\bibitem{chvatal1977aggregations}
V.~Chv{\'a}tal and P.~Hammer, ``Aggregations of inequalities,'' {\em Studies in
  Integer Programming, Annals of Discrete Mathematics}, vol.~1, pp.~145--162,
  1977.

\bibitem{tropp2007signal}
J.~Tropp and A.~Gilbert, ``Signal recovery from random measurements via
  orthogonal matching pursuit,'' {\em IEEE Trans. on info. theory}, 2007.

\bibitem{prechelt1998early}
L.~Prechelt, ``Early stopping-but when?,'' in {\em Neural Networks: Tricks of
  the trade}, pp.~55--69, Springer, 1998.

\bibitem{yao2007early}
Y.~Yao, L.~Rosasco, and A.~Caponnetto, ``On early stopping in gradient descent
  learning,'' {\em Constructive Approximation}, vol.~26, no.~2, pp.~289--315,
  2007.

\bibitem{raskutti2014early}
G.~Raskutti, M.~J. Wainwright, and B.~Yu, ``Early stopping and non-parametric
  regression: an optimal data-dependent stopping rule,'' {\em The Journal of
  Machine Learning Research}, vol.~15, no.~1, pp.~335--366, 2014.

\bibitem{mahsereci2017early}
M.~Mahsereci, L.~Balles, C.~Lassner, and P.~Hennig, ``Early stopping without a
  validation set,'' {\em arXiv preprint arXiv:1703.09580}, 2017.

\bibitem{duvenaud2016early}
D.~Duvenaud, D.~Maclaurin, and R.~Adams, ``Early stopping as nonparametric
  variational inference,'' in {\em Artificial Intelligence and Statistics},
  pp.~1070--1077, PMLR, 2016.

\bibitem{krizhevsky2009learning}
A.~Krizhevsky, G.~Hinton, {\em et~al.}, ``Learning multiple layers of features
  from tiny images,'' 2009.

\bibitem{kingma2015adam}
D.~P. Kingma and J.~L. Ba, ``Adam: A method for stochastic gradient descent,''
  in {\em ICLR: International Conference on Learning Representations},
  pp.~1--15, 2015.

\bibitem{srivastava2014dropout}
N.~Srivastava, G.~Hinton, A.~Krizhevsky, I.~Sutskever, and R.~Salakhutdinov,
  ``Dropout: a simple way to prevent neural networks from overfitting,'' {\em
  The journal of machine learning research}, vol.~15, no.~1, pp.~1929--1958,
  2014.

\bibitem{terry2021statistically}
J.~K. Terry, M.~Jayakumar, and K.~De~Alwis, ``Statistically significant
  stopping of neural network training,'' {\em arXiv preprint arXiv:2103.01205},
  2021.

\end{thebibliography}
\end{document}